\documentclass{article}

\usepackage{arxiv}

\usepackage[utf8]{inputenc} 
\usepackage[T1]{fontenc}    
\usepackage{hyperref}       
\usepackage{url}            
\usepackage{booktabs}       
\usepackage{nicefrac}       
\usepackage{microtype}      
\usepackage{lipsum}
\usepackage{fancyhdr}       
\usepackage{graphicx}       
\graphicspath{{media/}}     
\usepackage{float}
\usepackage{amsmath,amssymb,amsfonts,amsthm,bbm}
\usepackage{textcomp}
\usepackage{comment}
\usepackage{mathtools}
\usepackage[round]{natbib}

\title{Tensor-Fused Multi-View Graph Contrastive Learning}

\author{
  Yujia Wu \\
  Center for Data Science \\
  New York University \\
  \texttt{yw7702@nyu.edu} \\
   \And
  Junyi Mo \\
  Stern School of Business \\
  New York University \\
  \texttt{jm9847@stern.nyu.edu} \\
   \And
  Elynn Chen \\
  Stern School of Business \\
  New York University \\
  \texttt{elynn.chen@stern.nyu.edu} \\
   \And
  Yuzhou Chen \\
  Department of Statistics \\
  University of California, Riverside \\
  \texttt{yuzhou.chen@ucr.edu} \\
}
\providecommand{\angles}[1]{\left\langle #1 \right\rangle}
\providecommand{\paran}[1]{\left( #1 \right)}

\providecommand{\defeq}{:=}

\begin{document}
\maketitle

\begin{abstract}
Graph contrastive learning (GCL) has emerged as a promising approach to enhance graph neural networks' (GNNs) ability to learn rich representations from unlabeled graph-structured data. However, current GCL models face challenges with computational demands and limited feature utilization, often relying only on basic graph properties like node degrees and edge attributes. This constrains their capacity to fully capture the complex topological characteristics of real-world phenomena represented by graphs. To address these limitations, we propose Tensor-Fused Multi-View Graph Contrastive Learning (TensorMV-GCL), a novel framework that integrates extended persistent homology (EPH) with GCL representations and facilitates multi-scale feature extraction. Our approach uniquely employs tensor aggregation and compression to fuse information from graph and topological features obtained from multiple augmented views of the same graph. By incorporating tensor concatenation and contraction modules, we reduce computational overhead by separating feature tensor aggregation and transformation. Furthermore, we enhance the quality of learned topological features and model robustness through noise injected EPH. Experiments on molecular, bioinformatic, and social network datasets demonstrate TensorMV-GCL's superiority, outperforming 15 state-of-the-art methods in graph classification tasks across 9 out of 11 benchmarks, while achieving comparable results on the remaining two.  
The code for this paper is publicly available at \url{https://github.com/CS-SAIL/Tensor-MV-GCL.git}.
\end{abstract}


\section{Introduction}
In recent years, graph neural networks (GNNs) have become indispensable tools for learning from graph-structured data prevalent in various domains such as social networks \citep{davies2022realistic, fan2019graph}, molecular biology \citep{xia2023mole, sun2021mocl, wang2022molecular}, and recommendation systems 
\citep{di2021gnn, yang2023}. GNNs extend traditional neural networks by leveraging the relationships between nodes in a graph, using neighborhood aggregation to capture both node features and local graph structures. This method has proven effective in tasks such as node classification \citep{kipf2016semi, velivckovic2017graph}, link prediction \citep{wang2021benchmarking, ai2022structure, chen2022bscnets}, and graph classification \citep{wen2024tensor, ying2018hierarchical}. However, GNNs often rely heavily on task-specific labels, which can be scarce or expensive to obtain in many real-world applications. This dependence on labeled data limits the generalization ability of GNNs, especially when applied to noisy or sparse graphs. To mitigate this issue, graph contrastive learning (GCL) has emerged as a promising solution in the field of unsupervised and semi-supervised learning \citep{you2020graph, kefato2021self}. GCL enables models to learn meaningful representations by contrasting different augmented views of the same graph, thus reducing the need for extensive labeled data. By applying augmentation techniques such as node dropping, edge perturbation, and subgraph sampling, GCL encourages the model to capture invariant structural features, improving robustness and expressiveness in the face of data sparsity or noise.

Despite the progress made by GCL, existing frameworks continue to face several challenges. A key limitation is their inability to capture higher-order structural relationships and latent topological information embedded in graphs. Most GCL models focus on local node-level or edge-level augmentations, which often overlook the global topological properties necessary for a complete understanding of graph data. This limitation is particularly evident in tasks where capturing complex, multi-scale topological features is crucial, such as molecular interactions or long-range dependencies in network systems.  To overcome these issues, we propose the use of extended persistent homology (EPH), an advanced technique built on the foundations of persistent homology (PH) from topological data analysis (TDA). While PH has been studied by many articles \citep{horn2021topological, wasserman2018topological, carlsson2020topological}, it is often limited to single filtrations, reducing its ability to capture multi-dimensional structures in complex graphs. EPH introduces multi-dimensional filtrations, offering a richer, more flexible representation of both local and global topological information, making it more effective for understanding complex topological structures. Several applications of EPH have shown promising results in tasks, including link prediction, node classification, and graph classification \citep{zhao2020persistence, yan2021link, carriere2020perslay}. Among them, TopoGCL \citep{chen2024topogcl} is the first one to integrate EPH with GCL to capture important topological and geometric features and enhance representation learning. While TopoGCL shows notable promise in representation learning, it fails to (1) preserve the low-rank structure of representation tensors during learning, leading to inefficiencies and higher computational costs as the graph complexity increases, and (2) mitigate the additional noise introduced by EPH, which arises from its broader and more complex filtration process, potentially capturing irrelevant topological features and amplifying noise. 

In this paper, we address these issues by introducing a novel framework called Tensor-Fused Multi-View Graph Contrastive Learning (TensorMV-GCL), which innovatively combines tensor learning with graph contrastive learning.
Specifically, our framework comprises multiple channels for information aggregation: one for structural information and another for topological features. In the first channel, augmented graph views are processed through a shared-weight graph convolutional network (GCN) to learn structural representations.
In the second channel, we aggregate multiple extended persistent images (EPIs) in tensor form to capture multi-modal topological features.
These EPI tensors are injected with noise to counteract the noise introduced by the EPH procedure before being passed through a convolutional neural network (CNN) to further extract tensor representations.
Both channels incorporate a tensor concatenation layer for information aggregation and a tensor contraction layer for information compression, reducing redundancy and computational complexity.
Finally, all channels are integrated using a contrastive loss that fuses the structural and topological information, enabling our model to learn robust, comprehensive graph representations. By employing these techniques, we achieve more expressive and robust graph representations that better capture the underlying structural and topological properties of graphs. Extensive experiments validate the robustness and generalization capabilities of TensorMV-GCL, positioning it as a strong contender among state-of-the-art methods in graph classification tasks.

In brief, our key contributions are as follows:

\begin{itemize}
    \item TensorMV-GCL pioneers the integration of tensor learning with graph contrastive learning. Graph features from multiple views are aggregated in tensor forms, distinguishing different modes. Tensor contraction layers adaptively extract core features, reducing information redundancy and computational complexity. The synergy between tensor concatenation and contraction layers effectively extracts comprehensive graph representations from multiple views.
    \item We introduce a simple yet effective method to mitigate the noise introduced by the extended persistent homology's broad filtration process, thereby stabilizing topological representations and enhancing robustness.
    \item Our extensive experimental results demonstrate TensorMV-GCL's superior performance in graph classification tasks. It outperforms 15 state-of-the-art methods on 9 out of 11 datasets and achieves comparable performance on the remaining two.
\end{itemize}

\section{Related Work}
\textbf{Graph Neural Networks.} 
Graph Neural Networks (GNNs) are widely used for learning from graph-structured data by aggregating information from node neighborhoods, which enables effective representation learning for various tasks such as node classification \citep{velivckovic2017graph, zhang2019hyper}, link prediction \citep{kipf2016semi, wang2021benchmarking}, and graph classification \citep{nguyen2022universal, knyazev2019understanding}. A notable GNN variant is the Graph Convolutional Network (GCN), introduced by \cite{kipf2016semi}, which leverages convolution operations on graphs. GCNs propagate node information through graph edges, aggregating features from neighboring nodes to refine the representation of each node. This method effectively generalizes the convolutional operation from traditional grid-based data (such as images) to non-Euclidean graph data. However, GCNs, and GNNs more generally, tend to focus on simple node and edge features, limiting their ability to capture more complex topological structures in graphs. To overcome this limitation, recent approaches have explored integrating richer topological and structural information to enhance the expressive power of GNNs, thereby enabling models to better capture the intricate dependencies found in real-world graph data. Approaches like Graph Attention Networks \citep{GAT} and Graph Isomorphism Networks \citep{GIN}are examples of how GNNs continue to evolve to better capture these complexities, although challenges remain in balancing computational efficiency and capturing both local and global graph structures.

\noindent\textbf{Graph Contrastive Learning.} 
Graph Contrastive Learning (GCL) is a recent and powerful paradigm for self-supervised learning on graph-structured data. Self-supervised learning allows models to extract informative embeddings from vast amounts of unlabeled data by defining surrogate tasks that generate supervision signals from the data itself. In contrastive learning, the model aims to distinguish between similar and dissimilar data points, typically by maximizing the similarity between different augmented views of the same input (positive pairs) while minimizing the similarity to negative samples. Existing graph contrastive learning frameworks employ various contrasting modes, including global-global, global-local, and local-local contrastive learning. DGI \citep{velickovic2019deep} and InfoGraph \citep{sun2019infograph} both adopt global-local contrastive approaches, where global graph representations are contrasted with local node-level embeddings to maximize mutual information. Similarly, GraphCL \citep{you2020graph} and MolCLR \citep{wang2022molecular} use a global-global contrastive approach by contrasting two augmented views of the entire graph to enhance graph representations. Another mode, i.e. local-local approach \citep{jiao2020sub, zhu2021graph} is also applied by many models. For instance, SubG-Con \citep{jiao2020sub} proposes contrasting sampled subgraphs to capture localized patterns, which is effective on node-level tasks. Although GCL has shown remarkable success, existing frameworks often focus primarily on structural graph properties and may overlook the rich topological features that are crucial in many real-world graphs. Addressing this gap, methods like our proposed TensorMV-GCL aim to integrate both structural and topological information, allowing for multi-scale feature extraction and ultimately leading to more robust and comprehensive graph representations.

\vspace{1ex}
\noindent\textbf{Tensor-based Neural Networks.}  
Tensor learning has seen rapid advancements across various domains, including statistical modeling, projection methods, and system identification \cite{chen2023statistical,lin2021projection,liu2022identification,chen2024semi,chen2024time,zhang2024computation,li2023mev,chen2024data,chen2024dynamic}. Notable advancements have emerged in areas such as community detection, constrained optimization, and transfer learning \cite{chen2023community,chen2019constrained,chen2020modeling,chen2022modeling,chen2022transfer}, as well as in reinforcement learning and distributed computing \cite{chen2024reinforcement,chen2024distributed,chen2024advancing}. Further developments in high-dimensional tensor classification, stochastic optimization, and statistical inference \cite{chen2024high1,chen2024high2,chen2025stochastic,xu2025statistical} have expanded the applicability of tensor-based methodologies in modern machine learning.

While neural networks with tensor inputs efficiently process high-dimensional data, most approaches use tensors for computation rather than statistical analysis. \citet{cohen2016expressive} showed that deep networks can be understood through hierarchical tensor factorizations. Tensor contraction layers \citep{kossaifi2017tcl} and regression layers \citep{kossaifi2020trl} further regularize models, reducing parameters while maintaining accuracy. Recent advancements like the Graph Tensor Network \citep{xu2023graph} and Tensor-view Topological Graph Neural Network \citep{wen2024tensor} offer new frameworks for deep learning on large, multi-dimensional data. Additionally, substantial progress has been made in uncertainty quantification and representation learning using tensor-based approaches \cite{wu2024tensor,wu2024conditional1,wu2024conditional2,kong2024teaformers}.

However, there remains a lack of in-depth research on tensor-input neural networks with contrastive learning. To our best knowledge, TensorMV-GCL is the first approach to bridge this knowledge gap.

\vspace{1ex}
\noindent\textbf{Extended Persistence Homology.} Extended Persistence Homology (EPH) extends the classical persistence homology framework by considering not only the birth and death of topological features but also their evolution through different filtration levels, offering a more comprehensive characterization of data topology. The pioneering work by  \cite{cohen2009extending} laid the foundation for this approach, demonstrating its utility in capturing essential features that persist across various scales. Subsequent research, such as \cite{dey2014computing}, further refined these concepts by introducing algorithms for computing extended persistence diagrams, which visualize the birth, death, and extension of features. Applications of EPH have shown promising results in machine learning, where it helps in understanding the intrinsic geometry of high-dimensional data \citep{zhao2020persistence, yan2021link}. Despite its advantages, the integration of EPH with other machine learning techniques, such as GNNs and GCL, is still in its early stages, presenting an exciting avenue for future research. Our work aims to address these gaps by developing novel frameworks that leverage EPH to enhance the topological feature extraction and representation learning capabilities in graph-based models.

\section{Preliminaries}
\textbf{Problem Setting.} We consider an attributed graph $\mathcal{G} = (\mathcal{V}, \mathcal{E}, \boldsymbol{X})$, where $\mathcal{V}$ is the set of nodes with $|\mathcal{V}| = N$, $\mathcal{E}$ is the set of edges, and $\boldsymbol{X} \in \mathbb{R}^{N \times F}$ is the node feature matrix where $F$ represents the dimensionality of node features. Let $\boldsymbol{A} \in \mathbb{R}^{N \times N}$ be the symmetric adjacency matrix, where each entry $a_{ij}$ equals $\omega_{ij}$ if nodes $i$ and $j$ are connected, and is $0$ otherwise. Then, $\omega_{ij}$ is the edge weight with $\omega_{ij} \equiv 1$ for unweighted graphs. Additionally, $\boldsymbol{D}$ denotes the degree matrix of $\boldsymbol{A}$, defined as $d_{ii} = \sum_j a_{ij}$.

\vspace{1ex}
\noindent\textbf{Extended Persistent Homology.} 
Persistent homology (PH) is a computational method in algebraic topology used to analyze the multi-scale topological features of data by tracking their evolution across different scales. Specifically, PH captures changes in topological properties, such as connected components, loops, and higher-dimensional holes, as a filtration parameter changes, allowing it to detect features that persist over a range of scales, which are indicative of significant structure in the data \citep{edelsbrunner2002topological, zomorodian2004computing,chen2022transfer}.  By using a multi-resolution approach, PH addresses the inherent limitations of classical homology and enables the extraction of latent shape properties of $\mathcal{G}$. The main approach involves selecting appropriate scale parameters $\gamma$ and analyzing how the shape of $\mathcal{G}$ evolves as $\mathcal{G}$ changes with respect to $\gamma$. Instead of studying $\mathcal{G}$ as a single object, we study a filtration $\mathcal{G}_{\gamma_1} \subseteq \ldots \subseteq \mathcal{G}_{\gamma_n} = \mathcal{G}$ where ${\gamma_1} < \cdots < {\gamma_n}$. We can then record the birth $b_i$ and death $d_i$ of each topological feature, representing the scale at which each feature first and last appears in the sublevel filtration. However, PH has limitations in retaining information about features that persist indefinitely, which results in the loss of crucial latent topological information for certain applications. To address these limitations, an alternative approach is to supplement the sublevel filtration with its superlevel counterpart, which is called extended persistent homology (EPH) \citep{chen2024topogcl, cohen2009extending, dey2014computing}. Opposite to the sublevel filtration, we record the scale at which each feature first and last appears in the superlevel filtration $\mathcal{G}^{\gamma_1} \supseteq \mathcal{G}^{\gamma_2} \supseteq \ldots \supseteq \mathcal{G}^{\gamma_n}$. Hence, we can simultaneously assess topological features that persist and reappear in both directions, thereby providing a more comprehensive representation of the underlying topological properties of the data. Then, the extended topological information can be summarized as a multiset in $\mathbb{R}$, called the Extended Persistence Diagram (EPD), which can be denoted as $\text{EPD} = \{(b_\rho, d_\rho) \in \mathbb{R}^2\}$.

\vspace{1ex}
\noindent\textbf{Tensor Concatenation and Contraction.} 
The tensor structures' benefits stem from their reduced dimensionality in each mode -- since vectorizing tensors {\it multiplicatively} increases dimensionality -- and their potential statistical and computational advantages when their inherent multi-linear operations are leveraged \citep{kolda2009tensor}.

In this paper, tensor concatenation refers to the process of combining multiple tensors along a new mode, while tensor contraction denotes the extraction of core tensor factors from a tensor decomposition.
We examine three common tensor decompositions, defined as follows: consider an $M$-th order tensor $\mathcal{X}$ of dimension $D_1 \times \cdots \times D_M$. 
If $\mathcal{X}$ assumes a (canonical) rank-$R$ \textit{CP low-rank} structure, then it can be expressed as
\begin{equation} \label{eqn:cp}
    \mathcal{X} = \sum_{r=1}^R c_r \, \mathbf{u}_{1r} \circ \mathbf{u}_{2r} \circ \cdots \mathbf{u}_{Mr},
\end{equation}
where $\circ$ denotes the outer product, $\mathbf{u}_{mr} \in \mathbb{R}^{D_m}$ and $\|\mathbf{u}_{mr}\|_2 = 1$ for all mode $m \in [M]$ and latent dimension $r \in [R]$.
Concatenating all $R$ vectors corresponding to a mode $m$, we have
$\mathbf{U}_m = [\mathbf{u}_{m1}, \cdots, \mathbf{u}_{mR}] \in \mathbb{R}^{D_m \times R}$ which is referred to as the loading matrix for mode $m \in [M]$. 

If $\mathcal{X}$ assumes a rank-$(R_1, \cdots, R_M)$ \textit{Tucker low-rank} structure, then it writes 
\begin{equation} \label{eqn:tucker}
    \mathcal{X} = \mathcal{C} \times_1 \mathbf{U}_1 \times_2 \cdots \times_M \mathbf{U}_M = \sum_{r_1=1}^{R_1} \cdots \sum_{r_M=1}^{R_M} c_{r_1\cdots r_M} (\mathbf{u}_{1 r_1} \circ \cdots \circ \mathbf{u}_{M r_M}),
\end{equation}
where $\mathbf{u}_{m r_m}$ are all $D_m$-dimensional vectors, and $c_{r_1\cdots r_M}$ are elements in the $R_1 \times \cdots \times R_D$-dimensional core tensor $\mathcal{C}$. 

\textit{Tensor Train (TT) low-rank} approximates a $D_1 \times \cdots \times D_M$ tensor $\mathcal{X}$ with a chain of products of third order \textit{core tensors} $\mathcal{C}_i$, $i\in [M]$, of dimension $R_{i-1} \times D_i \times R_i$.  
Specifically, each element of tensor $\mathcal{X}$ can be written as 
\begin{equation} \label{eqn:tt}
    x_{i_1, \cdots, i_M} = \mathbf{c}_{1,1,i_1,:}^\top
    \times \mathbf{c}_{2,:,i_2,:} \times \cdots \times \mathbf{c}_{M,:,i_M,:}
    \times \mathbf{c}_{M+1,:,1,1}, 
\end{equation}
where $\mathbf{c}_{m,:,i_m,:}$ is an $R_{m-1} \times R_m$ matrix for $m \in [M] \cup \{M+1\}$. 
The product of those matrices is a matrix of size $R_0 \times R_{M+1}$.
Letting $R_0 = 1$, the first core tensor $\mathcal{C}_1$ is of dimension $1 \times D_1 \times R_1$, which is actually a matrix and whose $i_1$-th slice of the middle dimension (i.e., $\mathbf{c}_{1,1,i_1,:}$) is actually a $R_1$ vector. 
To deal with the "boundary condition" at the end, we augmented the chain with an additional tensor $\mathcal{C}_{M+1}$ with $D_{M+1} = 1$ and $R_{M+1} = 1$ of dimension $R_M \times 1 \times 1$.  
So the last tensor can be treated as a vector of dimension $R_M$.

CP low-rank \eqref{eqn:cp} is a special case where the core tensor $\mathcal{C}$ has the same dimensions over all modes, that is $R_m = R$ for all $m\in[M]$, and is super-diagonal.
TT low-rank is a different kind of low-rank structure and it inherits advantages from both CP and Tucker decomposition.
Specifically, TT decomposition can compress tensors as significantly as CP decomposition, while its calculation is as stable as Tucker decomposition.
\section{Methodology}
\begin{figure*}[htpb!]
  \centering
  \includegraphics[width=\linewidth]{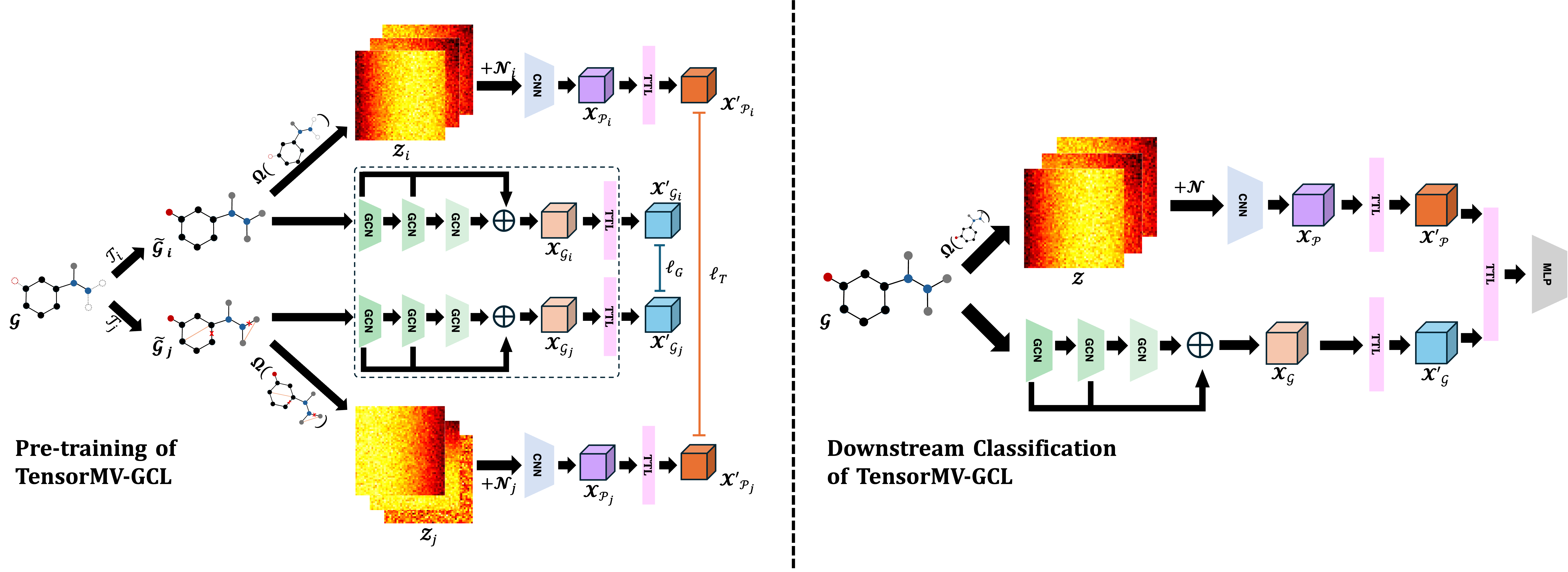}   
  \caption{In the pre-training stage (left), augmented graph views are processed through a shared GCN to generate node embeddings for Tensor-view Graph Contrastive Learning. In a parallel channel, a CNN is applied to extract features from EPH representations, with noise injected for robustness, then used for Stabilized Extended Persistent Images Contrastive Learning. Both channels adopt a TTL at the end to fuse the learned structural and topological features. In the downstream classification stage (right), a new graph is processed similarly, combining node embeddings with topological features extracted via the EPH-based CNN. The aggregated features pass through the TTL for final fusion, followed by an MLP for classification.}
  \label{figure1}
\end{figure*}

This section introduces our Tensor-Fused Multi-View Graph Contrastive Learning framework, TensorMV-GCL, which is depicted in Figure \ref{figure1}. As shown, our method comprises two main components. The first is Tensor-view Graph Contrastive Learning, represented by the inner two channels. This component aims to align two global representations derived from the same graph but from two different augmented views. The second component is Stabilized Extended Persistent Images Contrastive Learning, represented by the outer two channels. This component aligns two Extended Persistent Images (EPIs) extracted from two augmented views of a single graph. We introduce noise to these EPIs to enhance the stability and robustness of their representations.

\subsection{Tensor-view Graph Contrastive Learning}
Our first contrastive learning module is Tensor-view Graph Contrastive Learning, designed to leverage both structural and feature representations of graphs through a contrastive learning framework. Given a set of graphs $\{\mathcal{G}_{1}, \mathcal{G}_{2}, \dots, \mathcal{G}_{\mathcal{N}}\}$, we define each graph $\mathcal{G}_i$ by its node set $\mathcal{V}_i$, edge set $\mathcal{E}_i$, and node feature matrix $\boldsymbol{X}_i$. We have $\mathcal{G}_{i} = (\mathcal{V}_{i}, \mathcal{E}_{i}, \boldsymbol{X}_{i})$.

In order to capture multiple perspectives of each graph, we apply graph data augmentations, resulting in two augmented versions of each graph: $\tilde{{\mathcal{G}}}_{i} = \mathcal{T}_{i}(\mathcal{G})$ and $\tilde{{\mathcal{G}}}_{j} = \mathcal{T}_{j}(\mathcal{G})$. The augmentations applied are carefully designed to preserve the core structure and essential properties of the graph, while introducing variations that simulate different real-world conditions or incomplete data. This helps the model learn representations that are invariant to such changes, leading to more robust performance in downstream tasks.
After the graph data augmentations, we can construct the corresponding symmetric adjacency matrix and the degree matrix
$\tilde{{\mathcal{G}}}_{i} $ with ${\boldsymbol{A}}_{i}$ and ${\boldsymbol{D}}_{i}$; $\tilde{{\mathcal{G}}}_{j} $ with ${\boldsymbol{A}}_{j}$ and ${\boldsymbol{D}}_{j}$. 

We used a shared-weight Graph Convolutional Network (GCN) to encode these two augmented graphs. Specifically, the graph convolution operation involves multiplying the input of each layer by the $\tau$-th power of the normalized adjacency matrix. This $\tau$-th power encodes information from the $\tau$-th step of a random walk on the graph, allowing nodes to indirectly gather information from more distant nodes within the graph. Combined with an extra MLP layer, the representation learned at the $\ell$-th layer is
\begin{equation}
    \boldsymbol{C}_{{\mathcal{G}}_{i}}^{(\ell+1)} = f_{\text{MLP}}\left( \sigma \left( \widehat{\boldsymbol{A}}_{i}^\tau \boldsymbol{C}_{{\mathcal{G}}_{i}}^{(\ell)} \boldsymbol{W}^{(\ell)} \right) \right),
\end{equation}
where $\widehat{\boldsymbol{A}}_{i} = \tilde{\boldsymbol{D}}_{i}^{
-\frac{1}{2}} \tilde{\boldsymbol{A}}_{i} \tilde{\boldsymbol{D}}_{i}^{\frac{1}{2}}$, $\tilde{\boldsymbol{A}}_{i} = {\boldsymbol{A}}_{i} + \boldsymbol{I}$, and $\tilde{\boldsymbol{D}}$ is the degree matrix of $\tilde{\boldsymbol{A}}_{i}$, $\boldsymbol{C}_{\mathcal{G}_i}^{(0)} = \boldsymbol{X}_{i}$. 
$\sigma(\cdot)$ is a non-linear activation function. $\boldsymbol{W}^{(\ell)}$ is a trainable weight of $\ell$-th layer. $f_{\text{MLP}}$ is a MLP layer with batch normalization.

We concatenate all layers of $L$-layer graph convolutions $[\boldsymbol{C}_{{\mathcal{G}}_{i}}^{(1)}, \boldsymbol{C}_{{\mathcal{G}}_{i}}^{(2)}, \dots, \boldsymbol{C}_{{\mathcal{G}}_{i}}^{(L)}]$ to form a node embedding tensor $\boldsymbol{\mathcal{X}}_{\mathcal{G}_{i}}$ of dimension $N \times L \times D \times D$. Then, $\boldsymbol{\mathcal{X}}_{\mathcal{G}_{i}}$ is fed into a Tensor Transformation Layer (TTL) as $\boldsymbol{\mathcal{H}}_i^{(0)} = \boldsymbol{\mathcal{X}}_{\mathcal{G}_{i}}$. The dimension of $\boldsymbol{\mathcal{H}}_i^{(0)}$ is also $N \times L \times D \times D$. The output of TTL is $\boldsymbol{\mathcal{X}}'_{\mathcal{G}_{i}}$. Let \(\mathcal{X}'_{\mathcal{G}_i}\) denote the flattened version of \(\boldsymbol{\mathcal{X}}'_{\mathcal{G}_i}\), where \(\mathcal{X}'_{\mathcal{G}_i} \in \mathbb{R}^{N \times (L \cdot D \cdot D)}\). At the end of graph representation extraction, we have two embeddings corresponding to two different augmented graphs: $\mathcal{X}'_{\mathcal{G}_i}$ and $\mathcal{X}'_{\mathcal{G}_j}$

The core of the contrastive learning framework is to encourage the representations of two augmented views of the same graph to be similar while pushing apart representations of different graphs. To achieve this, we define the contrastive loss as:
\begin{equation}\label{tensorLoss}
    {\ell}_{i,G} = -\log\frac{\exp{(\text{sim}({\mathcal{X}}'_{\mathcal{G}_{i}}}, {\mathcal{X}}'_{\mathcal{G}_{j}})/\zeta)}{\sum\limits_{n=1, n \neq i,j}^{2\mathcal{N}} \exp(\text{sim}({\mathcal{X}}'_{\mathcal{G}_{i}}, {\mathcal{X}}'_{\mathcal{G}_{n}})/\zeta)}
\end{equation}
where $\text{sim}({\mathcal{X}}'_{\mathcal{G}_{i}}, {\mathcal{X}}'_{\mathcal{G}_{j}}) = {({\mathcal{X}}'_{\mathcal{G}_{i}})^\top {\mathcal{X}}'_{\mathcal{G}_{j}}} / {\|{\mathcal{X}}'_{\mathcal{G}_{i}}\| \, \|{\mathcal{X}}'_{\mathcal{G}_{j}}\|}$ represent the cosine similarity between two graph representations, and $\mathcal{X}'_{\mathcal{G}_{n}}$ is the flattened decomposed node embedding tensor representation of $\mathcal{G}_{n}$. The temperature parameter $\zeta$ smooths the distribution of similarities, helping balance between strong and weak similarities.

\subsection{Stabilized Extend Persistent Images Contrastive Learning}
Our second contrastive learning module is the Stabilized Extended Persistent Images Contrastive Learning. To capture the underlying topological information of each graph, we use $K$ vertex filtration functions, defined as $f_i: \mathcal{V} \mapsto \mathbb{R} \, \text{ for } \, i \in \{1, \ldots, K\}$. Each filtration function $f_i$ reveals a specific topological structure of the graph at different levels of connectivity. For the two augmented graphs $\tilde{\mathcal{G}}_{i}$ and $\tilde{\mathcal{G}}_{j}$, we construct two sets of $M$ EPIs of resolution $P \times P$, with each EPI corresponding to a filtration function, denoted as $M_i$ and $M_j$. With these two sets of EPIs, we can combine them respectively to construct the topological representations of $\tilde{\mathcal{G}}_{i}$ and $\tilde{\mathcal{G}}_{j}$, which denote as $\boldsymbol{\mathcal{Z}_i}$ and $\boldsymbol{\mathcal{Z}_j}$ with dimension of $K \times M \times P \times P$. To enhance the robustness of the model, we add Gaussian noise to $\boldsymbol{\mathcal{Z}}_i$ and $\boldsymbol{\mathcal{Z}}_j$ before further processing. Specifically, we generate noise tensors $\boldsymbol{\mathcal{N}}_i$ and $\boldsymbol{\mathcal{N}}_j$ with the same shape as $\boldsymbol{\mathcal{Z}}_i$ and $\boldsymbol{\mathcal{Z}}_j$, where each element of $\boldsymbol{\mathcal{N}}_i$ and $\boldsymbol{\mathcal{N}}_j$ follows a Gaussian distribution with mean $0$ and variance $\sigma^2=1$. The noisy tensors are then defined as:
\begin{equation}
    \boldsymbol{\mathcal{Z}}'_i = \boldsymbol{\mathcal{Z}}_i + \boldsymbol{\mathcal{N}}_i, \quad \boldsymbol{\mathcal{Z}}'_j = \boldsymbol{\mathcal{Z}}_j + \boldsymbol{\mathcal{N}}_j,
\end{equation}

Next, we use a shared-weight Convolutional Neural Network (CNN) and a global pooling layer to encode the hidden representations of $\boldsymbol{\mathcal{Z}}'_i$ and $\boldsymbol{\mathcal{Z}}'_j$ into topological tensor representations:
\begin{equation}
    \boldsymbol{\mathcal{X}}_{\mathcal{P}_i} = 
    \begin{cases}
        \displaystyle f_{\text{CNN}}(\boldsymbol{\mathcal{Z}}'_{i}) & \text{if } |M| = 1, \\
        \displaystyle \Phi_{\text{POOL}}(f_{\text{CNN}}(\boldsymbol{\mathcal{Z}}'_{i})) & \text{if } |M| > 1,
    \end{cases}
\end{equation}
where $f_{\text{CNN}}$ is a CNN, and $\Phi_{\text{POOL}}$ is a pooling layer that preserves the input information in a fixed-size representation. 
We divide the encoding process into two cases: (1) if the set of EPIs contains only one image, we apply a CNN directly to extract the latent features of $\boldsymbol{\mathcal{Z}}'_i$; (2) if there are multiple EPIs in the set, we employ a global pooling layer to aggregate the latent features into a fixed-size representation, ensuring consistent information retention across different numbers of EPIs. Then, the topological tensor representations are fed into a TTL with $\boldsymbol{\mathcal{H}}^{(0)} = \boldsymbol{\mathcal{X}}_{\mathcal{P}_i}$. The output of TTL is denoted as $\boldsymbol{\mathcal{X}}'_{\mathcal{P}_i}$. Let \(\mathcal{X}'_{\mathcal{P}_i}\) denote the flattened version of \(\boldsymbol{\mathcal{X}}'_{\mathcal{P}_i}\).
At the end of topological representation extraction, we have two flattened topological tensor representations corresponding to two augmented graphs: \(\mathcal{X}'_{\mathcal{P}_i}\) and \(\mathcal{X}'_{\mathcal{P}_j}\). Finally, we define the contrastive loss between these topological tensor representations to be:
\begin{equation}\label{EPHloss}
    \ell_{i,T} = -\log \frac{\exp \left( \text{sim}({\mathcal{X}}'_{\mathcal{P}_{i}}, {\mathcal{X}}'_{\mathcal{P}_{j}}) / \zeta \right)}{\sum\limits_{\substack{n=1, n \neq i,j}}^{2\mathcal{N}} \exp \left( \text{sim}({\mathcal{X}}'_{\mathcal{P}_{i}}, {\mathcal{X}}'_{\mathcal{P}_{n}}) / \zeta \right)},
\end{equation}
 where $\text{sim}({\mathcal{X}}'_{\mathcal{P}_{i}}, {\mathcal{X}}'_{\mathcal{P}_{j}}) ={({\mathcal{X}}'_{\mathcal{P}_{i}})^\top {\mathcal{X}}'_{\mathcal{P}_{j}}} / {\|{\mathcal{X}}'_{\mathcal{P}_{i}}\| \|{\mathcal{X}}'_{\mathcal{P}_{j}}\|}$, $\mathcal{X}'_{\mathcal{P}_{n}}$ is the flattened decomposed topological tensor representation of $\mathcal{G}_{n}$, and $\zeta$ is the temperature parameter. \\
 
 The final training objective function $\ell$ combines Equations (\ref{tensorLoss}) and (\ref{EPHloss}):
 \begin{equation}\label{totalloss}
     \ell = \alpha \times \sum\limits_{i=1}^{\mathcal{N}} {\ell}_{i,G} + \beta \times \sum\limits_{i=1}^{\mathcal{N}} {\ell}_{i,T},
 \end{equation}
 where $\alpha$ and $\beta$ are hyperparameters that balance the contribution of two contrastive losses. The default hyperparameters used for training are: $\alpha = 1$ and $\beta = 0.3$.

\subsection{Tensor Transformation Layer}
The \textit{Tensor Transformation Layer (TTL)} preserves the tensor structure of the feature matrix $\boldsymbol{{X}}$, which has dimensions given by $D = \prod_{m=1}^{M} D_m$, while maintaining hidden representations throughout the network. Let $\mathcal{L}$ be a positive integer, and let $\mathbf{a} = [a^{(1)}, \ldots, a^{(\mathcal{L}+1)}]$  denote the width of each layer. A deep ReLU Tensor Neural Network can be expressed as a function mapping in the following form:
\begin{equation}
    f(\boldsymbol{\mathcal{X}}) = \mathcal{L}^{(L+1)} \circ \sigma \circ \mathcal{L}^{(L)} \circ \sigma \circ \ldots \circ \sigma \circ \mathcal{L}^{(1)}(\boldsymbol{{X}}),
\end{equation}
where $\sigma(\cdot)$ represents an element-wise activation function. The affine transformation $\mathcal{L}^{(\ell)}(\cdot)$, along with the hidden input and output tensors at the $\ell$-th layer, denoted by $\boldsymbol{\mathcal{H}}^{(\ell+1)}$ and $\boldsymbol{\mathcal{H}}^{(\ell)}$, are defined as follows:
\begin{equation}
\begin{aligned}
\mathcal{L}^{(\ell)}\paran{\boldsymbol{\mathcal{H}}^{(\ell)}} \defeq \angles{\boldsymbol{\mathcal{W}}^{(\ell)}, \boldsymbol{\mathcal{H}}^{(\ell)}} + \boldsymbol{\mathcal{B}}^{(\ell)}, \\
    \text{and}\quad
    \boldsymbol{\mathcal{H}}^{(\ell+1)} \defeq \sigma\paran{\mathcal{L}^{(\ell)}\paran{\boldsymbol{\mathcal{H}}^{(\ell)}}},
\end{aligned}
\end{equation}
where
$\boldsymbol{\mathcal{H}}^{(0)} = \boldsymbol{\mathcal{X}}$ takes the tensor feature, 
$\angles{\cdot, \cdot}$ is the tensor inner product, 
and {\em low-rank weight} tensor $\boldsymbol{\mathcal{W}}^{(\ell)}$ and a bias tensor $\boldsymbol{\mathcal{B}}^{(\ell)}$. 
The tensor structure kicks in when we incorporate tensor low-rank structures such as {\it CP low-rank} \eqref{eqn:cp}, {\it Tucker low-rank} \eqref{eqn:tucker}, and {\it Tensor Train low-rank} \eqref{eqn:tt}. We focus on {\it CP low-rank} in this paper due to its superior performance \citep{wen2024tensor}. 
\section{Experiments}

\subsection{Experiment Settings}

\textbf{Datasets.} 
In this work, we evaluate the performance of our TensorMV-GCL model on graph classification tasks on chemical compounds, molecule compounds, and social networks datasets. For chemical compounds, the datasets used include NCI1, DHFR, MUTAG, BZR, and COX2~\citep{sutherland2003spline,kriege2012subgraph}, which are composed of graphs representing chemical compounds with nodes as atoms and edges as chemical bonds. In the case of molecules compounds, the datasets employed are PROTEINS, D\&D, PTC\_MR, and PTC\_FM~\citep{helma2001predictive,dobson2003distinguishing,borgwardt2005protein}, where each protein is depicted as a graph with nodes representing amino acids and edges denoting interactions like physical bonds or spatial proximity. We also include a social network dataset, IMDB-B, in which nodes represent actors and actresses, with edges connecting those who have appeared in the same movie. Each dataset has graphs belonging to a certain class and our model aims to classify graphs' classes. Table \ref{tab:datasets} summarizes the
characteristics of all eleven datasets used in our experiments.
\begin{table}[h!]
\centering
\caption{Summary statistics of the benchmark datasets.\label{tab:datasets}}
\resizebox{0.45\columnwidth}{!}{
\begin{tabular}{lcccc}
\toprule
\textbf{{Dataset}} & \textbf{{\# Graphs}} &\textbf{{Avg.} $|\mathcal{V}|$} & \textbf{{Avg.} $|\mathcal{E}|$} & \textbf{{\# Class}} \\
\midrule
NCI1 & 4110 & 29.87 & 32.30 & 2\\
PROTEINS &1113 &39.06 &72.82 &2 \\
DD &1178 &284.32 &715.66 &2 \\
MUTAG &188 &17.93 &19.79 &2 \\
DHFR & 756 & 42.43& 44.54 & 2\\
BZR &405 &35.75 &38.36 &2 \\
COX2 &467 &41.22 &43.45 &2 \\
PTC\_MR &  344 & 14.29 & 14.69 & 2\\
PTC\_FM &  349 & 14.11 & 14.48 & 2\\
IMDB-B & 1000 & 19.77 & 96.53 & 2\\
REDDIT-B & 2000 & 429.63 & 497.75 & 2 \\
\bottomrule
\end{tabular}}
\end{table}
\begin{table}[htpb!]
    \centering
    \caption{Optimal Augmentations of Each Dataset}
    \label{augmentation}
    \resizebox{0.3\textwidth}{!}{ 
    \begin{tabular}{l c c}
    \toprule
    \textbf{Dataset} & \textbf{Augmentations} \\ 
    \midrule
    \text{NCI1} & NodeDrop + AttrMask \\
    \text{PROTEINS} & NodeDrop + EdgePert \\
    \text{DD} & NodeDrop + Subgraph \\
    \text{MUTAG} & NodeDrop + EdgePert \\
    \text{DHFR} & EdgePert + Identical \\
    \text{BZR} & NodeDrop + AttrMask \\
    \text{COX2} & NodeDrop + Subgraph \\
    \text{PTC\_MR} & NodeDrop + EdgePert \\
    \text{PTC\_FM} & NodeDrop + EdgePert \\
    \text{IMDB-B} & Subgraph + Identical \\
    \text{REDDIT-B} & Subgraph + Identical \\ 
    \bottomrule
    \end{tabular}%
    }
\end{table}
\vspace{1ex}

\noindent\textbf{Baseline.} 
We compare our proposed TensorMV-GCL on 11 real-world datasets with 15 state-of-the-art baselines including (1) Graphlet Kernel (GL)~\citep{KL}, (2) Contrastive Multi-View Representation Learning on Graphs (MVGRL)~\citep{MVGRL}, (3) Weisfeiler-Lehman Sub-tree Kernel (WL)~\citep{WL}, (4) Deep Graph Kernels (DGK)~\citep{DGK}, (5) node2vec ~\citep{node2vec}, (6) sub2vec~\citep{sub2vec}, (7) graph2vec ~\citep{graph2vec}, (8) InfoGraph~\citep{infoGraph}, (9) Graph Contrastive Learning with Augmentations (GraphCL)~\citep{GraphCL}, (10) Graph Contrastive Learning Automated (JOAO)~\citep{JOAO}, (11) Adversarial Graph Augmentation to Improve Graph Contrastive Learning (AD-GCL)~\citep{ADGCL}, (12) AutoGCL~\citep{autoGCL}, (13) A Review-aware Graph Contrastive Learning Framework for Recommendation (RGCL)~\citep{Shuai_2022}, (14) GCL-TAGS~\citep{lin2022spectrum}, and (15) TopoGCL~\citep{chen2024topogcl}.

\vspace{1ex}
\noindent\textbf{Training Setup.}
We run our experiments on a single NVIDIA Quadro RTX 8000 GPU card, which has up to 48GB of memory. To train the end-to-end TensorMV-GCL model, we use the Adam optimizer with a learning rate of 0.001.  We use ReLU as the activation function $\sigma(\cdot)$ across our model. For the resolution of the $\text{EPI}$, we set the size to be $P = 50$. In our experiments, we consider $\mathcal{K} = 4$ different filtrations, i.e., degree-based, betweenness-based, closeness-based, and eigenvector-based filtrations. Depending on the dataset, we set batch sizes of either 16 or 32. The optimal number of hidden units for each layer in the graph convolution and MLPs are explored from the search space $\{16, 32, 64, 128, 256\}$. The default tensor decomposition method of TTL is CP low-rank decomposition mentioned in equation \ref{eqn:cp}. The number of hidden units of TTL is 32. In addition, our model has 3 layers in the graph convolution blocks and 2 layers in the MLPs, with a dropout rate of 0.5 for all datasets. For the contrastive pre-training stage, we train our model for up to 350 epochs to make sure it is fully learned. According to different datasets, we apply different augmentations as shown in Table \ref{augmentation}.
\begin{table*}[!h]
    \centering
    \caption{Classification Performance (Accuracy in \%) comparison across different models on various datasets (Best results are in \textbf{bold}).}
    \label{table1}
    \resizebox{\textwidth}{!}{%
    \begin{tabular}{l c c c c c c c c c c c}
    \toprule
    \textbf{Model}      & \textbf{NCI1}        & \textbf{PROTEINS}    & \textbf{DD}           & \textbf{MUTAG}       & \textbf{DHFR}        & \textbf{BZR}         & \textbf{COX2}        & \textbf{PTC\_MR}     & \textbf{PTC\_FM}  & \textbf{IMDB-B} & \textbf{REDDIT-B} \\ 
    \midrule
    GL         & N/A         & N/A         & N/A          & 81.66±2.11  & N/A         & N/A         & N/A         & 57.30±1.40 & N/A      & 65.87±0.98 & 77.34±0.18 \\ 
    MVGRL      & N/A & N/A & N/A & 75.40±7.80 & N/A & N/A & N/A & N/A & N/A & 63.60±4.20 & 82.00±1.10 \\
    WL         & 80.01±0.50  & 72.92±0.56  & 74.00±2.20   & 80.72±3.00  & N/A         & N/A         & N/A         & 58.00±0.50 & N/A      & 72.30±3.44 & 68.82±0.41 \\ 
    DGK        & 80.31±0.46  & 73.30±0.82  & N/A          & 87.44±2.72  & N/A         & N/A         & N/A         & 60.10±2.60 & N/A      & 66.96±0.56 & 78.04±0.39 \\ 
    node2vec   & 54.89±1.61  & 57.49±3.57  & N/A          & 72.63±10.20 & N/A         & N/A         & N/A         & N/A        & N/A      & 56.40±2.80 & 69.70±4.10 \\ 
    sub2vec    & 52.84±1.47  & 53.03±5.55  & N/A          & 61.05±15.80 & N/A         & N/A         & N/A         & N/A        & N/A      & 55.26±1.54 & 71.48±0.41 \\ 
    graph2vec  & 73.22±1.81  & 73.30±2.05  & N/A          & 83.15±9.25  & N/A         & N/A         & N/A         & N/A        & N/A      & 71.10±0.54 & 75.78±1.03 \\ 
    InfoGraph  & 76.20±1.06  & 74.44±0.31  & 72.85±1.78   & 89.01±1.13  & 80.48±1.34  & 84.84±0.86  & 80.55±0.51  & 61.70±1.40 & 61.55±0.92   & 73.03±0.87 & 82.50±1.42 \\ 
    GraphCL    & 77.87±0.41  & 74.39±0.45  & 78.62±0.40   & 86.80±1.34  & 68.81±4.15  & 84.20±0.86  & 81.10±0.82  & 61.30±2.10 & 65.26±0.59   & 71.14±0.44 & 89.53±0.84\\ 
    JOAO       & 78.07±0.47  & 74.55±0.41  & 77.32±0.54   & 87.35±1.02  & N/A         &
    N/A & N/A & N/A & N/A & 70.21±3.08 & 85.29±1.35\\
    AD-GCL     & 73.91±0.77  & 73.28±0.46  & 75.79±0.87   & 88.74±1.85  & 75.66±0.62  & 85.97±0.63  & 78.68±0.56  & 63.20±2.40 & 64.99±0.77   & 70.21±0.68 & 90.07±0.85\\ 
    AutoGCL    & \textbf{82.00±0.29}  & 75.80±0.36  & 77.57±0.60   & 88.64±1.08  & 77.33±0.76  & 86.27±0.71  & 79.31±0.70  & 63.10±2.30 & 63.62±0.55   & 72.32±0.93 & 88.58±1.49 \\ 
    RGCL       & 78.14±1.08  & 75.03±0.43  & 78.86±0.48   & 87.66±1.01  & 76.37±1.35  & 84.54±1.67  & 79.31±0.68  & 61.43±2.50 & 64.29±0.32   & 71.85±0.84  & 90.34±0.58\\ 
    GCL-TAGS   & 71.43±0.49  & 75.78±0.52  & N/A          & 89.12±0.76  & N/A         & N/A         & N/A         & N/A        & N/A          & 73.65±0.69  & 83.62±0.64\\
    TopoGCL    & 81.30±0.27  & 77.30±0.89  & 79.15±0.35 & 90.09±0.93 & 82.12±0.69 & 87.17±0.83 & 81.45±0.55 & 63.43±1.13 & 67.11±1.08   & 74.67±0.32 & \textbf{90.40±0.53}\\ \hline
    {\bf TensorMV-GCL (ours)}    & 74.63±1.78 & \textbf{80.92±2.05} & \textbf{89.65±2.23} & \textbf{98.89±0.59} & \textbf{82.61±3.79} & \textbf{94.56±2.25} & \textbf{89.58±2.26} & \textbf{87.81±3.19} & \textbf{89.08±6.58} & \textbf{76.79±2.56} & \text{90.27±0.85} \\ 
    \bottomrule
    \end{tabular}%
    }
\end{table*}

\begin{table*}[htpb!]
    \centering
    \caption{Ablation studies.}
    \label{ablation} 
    \resizebox{\textwidth}{!}{%
    \begin{tabular}{l c c c c c c c c c c c}
    \toprule
    \textbf{Architecture}      & \textbf{NCI1}        & \textbf{PROTEINS}           & \textbf{MUTAG}       & \textbf{DHFR}        & \textbf{BZR}         & \textbf{COX2}        & \textbf{PTC\_MR}     & \textbf{PTC\_FM}  & \textbf{IMDB-B}   \\ 
    \midrule
    TensorMV-GCL With PH   & 75.54±2.56  & 80.36±3.42  & 92.24±2.77 & 80.85±2.64 & 94.34±2.46  & 89.02±3.64     & 64.52±3.23         & 68.44±1.64        & 75.26±2.50
       \\
    TensorMV-GCL W/o TDA channel & N/A & 73.51±2.29 & 97.87±2.94  & 75.45±4.24 & 84.89±3.93 & 87.93±2.17 
    & 62.13±2.39  & 66.80±4.28 & N/A  \\
    TensorMV-GCL W/o noise    & \textbf{75.68±2.21}  & 78.73±2.57 & 96.25±3.59 & \textbf{83.44±4.02} & 91.13±2.41 & 84.59±4.21
    & 81.24±6.28 & 75.13±4.85
       & 72.44±3.05 \\ 
    

    TensorMV-GCL W/o TTL Layers & N/A & 72.38±1.43 & 96.74±3.23  & 78.21±4.64 &  89.75±1.20 & 88.39±1.97 & 63.53±4.60 & 78.43±2.76   & N/A  \\
    \midrule
    {\bf TensorMV-GCL (ours)}     & 74.63±1.78 & \textbf{80.92±2.05}  & \textbf{98.89±0.59} & 82.61±3.79 & \textbf{94.56±2.25} & \textbf{89.58±2.26} & \textbf{87.81±3.19} & \textbf{89.08±6.58} & \textbf{76.79±2.56} \\
    \bottomrule
    \end{tabular}%
    }
\end{table*}

Since we do not adopt a project head during the contrastive pre-training stage, we further apply MLPs as a classifier to perform our graph classification downstream tasks. The input for the classifier is the node embedding tensor $\mathcal{X}'_{\mathcal{G}}$ and the topological feature tensor $\mathcal{X}'_{\mathcal{P}}$ of the original graph. The detailed architecture is shown in Figure \ref{figure1}. We conduct end-to-end fine-tuning with the Adam optimizer with a learning rate of 0.001.  We use ReLU as the activation function $\sigma(\cdot)$ across our model. We train our end-to-end finetune for up to 100 epochs.

\subsection{Classification Performance}
As shown in Table \ref{table1}, we observe that TensorMV-GCL consistently outperforms the runner-up models across nearly all datasets. 

\vspace{1ex}
\noindent\textbf{Comparative Performance Analysis.} More specifically, we observe that TensorMV-GCL provides significant relative improvements over traditional graph embedding methods like node2Vec and Sub2Vec, with gains exceeding 10\% on most datasets. One key reason for this improvement is that, unlike traditional graph embedding methods, TensorMV-GCL incorporates both topological features and graph convolutional networks (GCNs). This allows the model to consider not only local node information but also the global structure of the entire graph. Traditional methods typically focus on node-level proximity and local neighborhoods, missing out on the richer, higher-order topological relationships that are essential for fully understanding complex graphs. By integrating topological features, TensorMV-GCL is able to capture a more comprehensive representation of both node-level and whole-graph information, leading to superior performance in graph classification tasks.

In addition, when compare to popular contrastive learning frameworks such as GraphCL, JOAO, and AD-GCL, TensorMV-GCL achieves up to a 9.5\% relative improvement. The advantage of our tensor-based contrastive learning approach lies in its ability to capture multi-scale structural and topological information from multiple augmented graph views. While traditional contrastive learning methods typically focus on contrasting graph views based on simple structural transformations, TensorMV-GCL goes further by leveraging topological insights through extended persistent homology (EPH). This allows the model to retain critical information at both local and global levels, improving its ability to distinguish between subtle variations in graph structures. 

\vspace{1ex}
\noindent\textbf{Scalability of TensorMV-GCL.}
As expected, our proposed TensorMV-GCL significantly outperforms recent state-of-the-art models such as AutoGCL, InfoGraph, and TopoGCL on most datasets. In particular, TensorMV-GCL demonstrates strong results in small dataset benchmarks such as MUTAG, BZR, COX2, and both PTC\_MR and PTC\_FM, which are often challenging due to their limited data sizes. The model’s ability to effectively capture both structural and topological features enables it to outperform these state-of-the-art approaches, especially in scenarios where extracting meaningful information from small datasets is critical. Additionally, to test TensorMV-GCL's scalability, we conducted experiments on datasets like NCI1, REDDIT-B, and DD, which either consist of a large number of graphs or feature graphs with significantly higher average node and edge counts. The results show that TensorMV-GCL is capable of maintaining high performance, even when applied to large-scale graphs. In summary, our experimental results demonstrate that TensorMV-GCL effectively captures both structural and topological information in graphs, offering a significant performance boost in a wide range of graph classification tasks.

\subsection{Ablation Studies}
To better understand the importance of different components in TensorMV-GCL, we have conducted 4 ablation studies during the pre-training stage. Specifically, we evaluated the impact of (1) using Persistent Homology (PH) instead of Extended Persistent Homology (EPH), (2) disabling the Stabilized Extended Persistent Images Contrastive Learning, i.e., set $\beta = 0$ in equation \ref{totalloss}, (3) removing the noise injection from EPH, and (4) omitting the Tensor Transformation Layer (TTL) at the end of all channels. We tested the ablations on the NCI1, PROTEINS, MUTAG, DHFR, BZR, COX2, PTC\_MR, PTC\_FM, and IMDB-B datasets. Results are shown in Table \ref{ablation}.

\vspace{1ex}
\noindent\textbf{The Effect of EPH.}
The first row presents the results using Persistent Homology (PH) alone, which leads to inferior performance on all datasets except NCI1. While switching from Extended Persistent Homology (EPH) to PH results in a performance drop, it is not substantial for most datasets, indicating that PH still captures some essential topological features. However, the comprehensive representation provided by EPH, which considers both sublevel and superlevel filtrations, generally results in higher performance across the majority of datasets.

As shown in the second row, the most significant performance impact occurs when we completely disable the Stabilized Extended Persistent Images Contrastive Learning, i.e., the TDA channel, meaning we do not use PH or EPH at all. In this case, the performance drops dramatically across all datasets, demonstrating the critical role of topological data analysis in the overall architecture. This highlights that, while both PH and EPH contribute meaningfully to feature extraction, entirely removing the topological learning channel severely limits the model's ability to capture key structural properties, leading to a substantial degradation in performance.

\vspace{1ex}
\noindent\textbf{The Effect of Adding Noise.}
In the third row, we evaluate our model without noise injection applied to the EPH-based features. While most datasets show a drop in performance compared to the configuration with noise injection, NCI1 and DHFR performed better without it. This suggests that although noise injection typically enhances robustness by helping the model learn more stable and generalized topological features, it may not be necessary for certain datasets where the model already generalizes effectively. Adding noise during the training process offers several advantages. It helps prevent overfitting by acting as a form of regularization, ensuring the model doesn't become overly specialized to the training data. Furthermore, noise injection introduces variability, which forces the model to learn more robust features that are less sensitive to minor perturbations in the data. This leads to improved generalization across different datasets. For larger datasets, we observe that the improvement from noise injection was either limited or slightly negative, likely because these datasets already provide sufficient diversity and variation to avoid overfitting. In contrast, for smaller datasets, the improvement was much more pronounced. Smaller datasets are more prone to overfitting, where noise injection serves as a crucial mechanism to encourage the model to focus on more generalized patterns, improving its overall performance.

\vspace{1ex}
\noindent\textbf{The Effect of Tensor Transformation Layer.}
In the fourth row, we evaluated the model after removing the Tensor Transformation Layer (TTL) from the end of each channel. The results show a noticeable drop in performance across all datasets, indicating that the TTL plays a crucial role in the success of the model. The TTL is essential for preserving the tensor structures generated during feature extraction and for effectively incorporating low-rankness, which helps reduce redundancy while retaining important information.

By maintaining the integrity of the tensor structure, the TTL ensures that the fused features from both the graph and topological channels remain coherent and meaningful, allowing the model to leverage both local and global information more effectively. Additionally, the incorporation of low-rankness enables the model to manage the complexity of high-dimensional tensor data, leading to better generalization and more robust feature learning. Without the TTL, the model struggles to process the rich information encoded in the tensors, resulting in a significant drop in overall performance.

The ablation studies highlight the importance of key components in TensorMV-GCL. EPH-based method with noise injection improves generalization and feature extraction, particularly in smaller datasets. Removing the Stabilized Extended Persistent Images Contrastive Learning (TDA channel) led to significant performance drops across all datasets, emphasizing its necessity. Similarly, the Tensor Transformation Layer (TTL) is crucial for preserving tensor structure and incorporating low-rankness, further enhancing performance. These elements together enable TensorMV-GCL to effectively capture both structural and topological features, ensuring robust performance across diverse graph tasks.

\section{Conclusion}
In this work, we introduce TensorMV-GCL, a novel model for graph classification that leverages tensor decomposition, Extended Persistent Homology (EPH), and noise augmentation to capture essential topological and structural features of graphs. Through extensive experiments across multiple datasets, we demonstrated that TensorMV-GCL consistently outperforms existing methods, showcasing its robustness and effectiveness in both small-scale and large-scale graph learning tasks. Our ablation study further emphasized the importance of EPH and noise addition, validating their roles in enhancing model performance and generalization.

Despite these promising results, our work has some limitations. First, while TensorMV-GCL achieves state-of-the-art performance in most cases, it may not fully capture complex, long-range dependencies between nodes due to the inherent limitations of current graph contrastive learning architectures. Additionally, the computational cost of EPH, especially on large-scale graphs, can be high, which could limit the scalability of our approach in certain applications.

For future work, we plan to explore the integration of graph transformers into the graph self-supervised learning framework. Transformers have shown remarkable success in capturing global dependencies in other domains, and we believe incorporating them into our framework could further improve performance. By combining tensor representations with the powerful attention mechanisms of graph transformers, we aim to enhance the model’s ability to capture both local and global graph features. This extension could lead to more scalable and effective models, pushing the boundaries of graph-based learning in a variety of domains.
\clearpage
\bibliography{bib/reference,bib/tensor}

\end{document}